\documentclass[10pt,twocolumn,letterpaper]{article}

\PassOptionsToPackage{table}{xcolor} 
\usepackage{cvpr}

\usepackage[dvipsnames]{xcolor}

\definecolor{cvprblue}{rgb}{0.21,0.49,0.74}
\usepackage[pagebackref,breaklinks,colorlinks,citecolor=cvprblue]{hyperref}
\usepackage{float} 
\usepackage{graphicx} 
\usepackage{dblfloatfix} 
\usepackage{csvsimple} 
\usepackage{booktabs}  
\usepackage{array}      
\usepackage{mathtools}
\usepackage{amsmath}
\usepackage{relsize} 
\usepackage[numbers, sort&compress]{natbib}
\usepackage{multirow}

\def\paperID{oIoNLgPNmX} 
\def\confName{3DV\xspace}
\def\confYear{2026\xspace}

\title{GOGS: High-Fidelity Geometry and Relighting for Glossy Objects \\via Gaussian Surfels}

\author{
  Xingyuan Yang\quad Min Wei \\ 
  Chengdu University of Information Technology \\ 
  {\tt\small yangcyanx@gmail.com}\quad{\tt\small weimin@cuit.edu.cn} 
}
\begin{document}
\twocolumn[{%
    \maketitle 
    \vspace{-2em} 
    \begin{figure}[H] 
        \hsize=\textwidth 
        \centering 
        \includegraphics[width=1\textwidth]{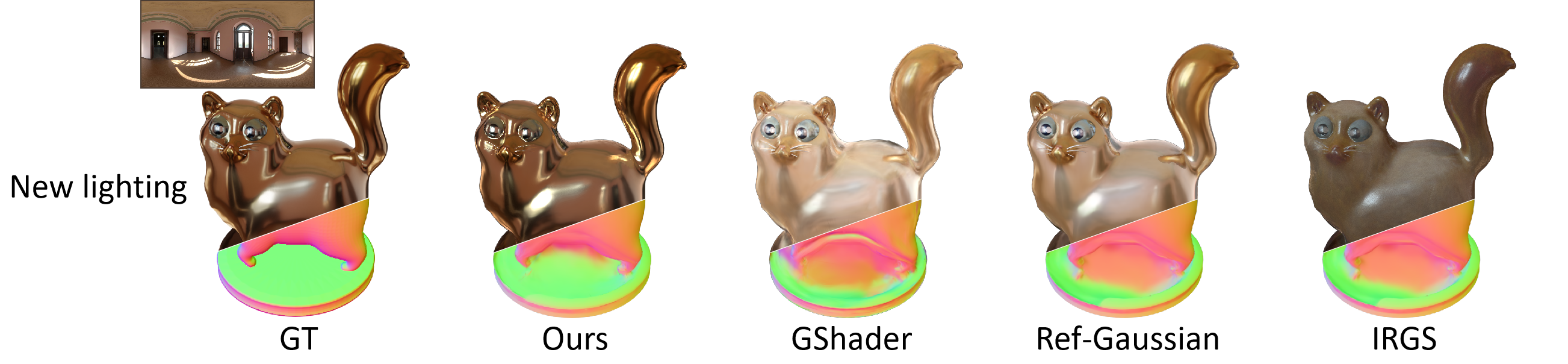} 
        \caption{Our method achieves accurate geometry recovery and photorealistic rendering under novel illumination. Our rendering closely matches the GT with realistic specular effects, outperforming competitors suffering from material errors or blurred specular. Normal maps accurately capture inter-reflection geometry (e.g. cat plate reflections), unlike competitors that fail under specular interference.}
        \label{fig:highlight}
    \end{figure}
}]

\setlength{\abovedisplayskip}{4pt}
\setlength{\belowdisplayskip}{4pt}
\begin{abstract}
\vspace{-1em}
Inverse rendering of glossy objects from RGB imagery remains fundamentally limited by inherent ambiguity. Although NeRF-based methods achieve high-fidelity reconstruction via dense-ray sampling, their computational cost is prohibitive. Recent 3D Gaussian Splatting achieves high reconstruction efficiency but exhibits limitations under specular reflections. Multi-view inconsistencies introduce high-frequency surface noise and structural artifacts, while simplified rendering equations obscure material properties, leading to implausible relighting results. To address these issues, we propose GOGS, a novel two-stage framework based on 2D Gaussian surfels. First, we establish robust surface reconstruction through physics-based rendering with split-sum approximation, enhanced by geometric priors from foundation models. Second, we perform material decomposition by leveraging Monte Carlo importance sampling of the full rendering equation, modeling indirect illumination via differentiable 2D Gaussian ray tracing and refining high-frequency specular details through spherical mipmap-based directional encoding that captures anisotropic highlights. Extensive experiments demonstrate state-of-the-art performance in geometry reconstruction, material separation, and photorealistic relighting under novel illuminations, outperforming existing inverse rendering approaches.
\end{abstract}    
\vspace{-1em}
\section{Introduction}
\label{sec:intro}
Inverse rendering of glossy objects poses formidable challenges in computer vision and graphics. These surfaces exhibit strong view-dependent effects, such as specular highlights, environmental reflections, and inter-reflections, that inherently undermine multi-view consistency. This fundamental conflict severely impedes the accurate recovery of geometry and material properties from multi-view imagery. Although methods~\cite{mildenhall2021nerf, wang2021neus, barron2021mipnerf, barron2022mipnerf360, muller2022instantngp, chen2022tensorf, verbin2022refnerf, liu2023nero, liang2023envidr, jin2023tensoir} based on Neural Radiance Fields (NeRF) demonstrate promise in modeling complex light transport, their reliance on dense ray sampling compromises computational efficiency for practical deployment.

Recent advances in 3D Gaussian Splatting (3DGS)~\cite{kerbl20233dgs} deliver efficient scene reconstruction through explicit representations and real-time rendering. However, subsequent methods exhibit various limitations: certain approaches~\cite{guedon2024sugar, yu2024mipsplatting, yu2024gof, chen2024pgsr, lu2024scaffold, huang20242dgs} exhibit limitations for glossy objects, while others achieve plausible geometry reconstruction for glossy surfaces, yet either fail to perform material decomposition~\cite{tang20243igs, yang2024specgs, bi2024gs3, zhang2025refgs} or persistently yield blurred decomposition results despite enabling inverse rendering~\cite{jiang2024gaussianshader, liang2024gsir, gao2024r3dg, ye20243dgsdr, yao2024refgaussian, gu2025irgs}.
Specifically, geometry reconstruction is severely compromised by view-dependent specular reflections that violate multi-view consistency. This manifests itself as both high-frequency surface noise and structural artifacts, particularly in inter-reflection regions. Simultaneously, material decomposition is impaired by a simplified rendering equation and neglected indirect illumination, leading to blurred albedo, metallic, and roughness estimates that yield physically implausible relighting results.

We propose GOGS, a novel geometry prior-guided framework based on 2D Gaussian surfels that resolves these limitations through a two-stage pipeline. First, we establish robust surface reconstruction through physics-based rendering with split-sum approximation, enhanced by geometric priors from foundation models to explicitly enforce curvature continuity and mitigate ambiguities under specular interference. Subsequently, we perform physically accurate material decomposition on this optimized geometry by evaluating the full rendering equation via Monte Carlo importance sampling, where visibility and indirect radiance are computed through differentiable 2D Gaussian ray tracing. To address fidelity limit in specular rendering, we further introduce a spherical mipmap-based specular compensation mechanism that adaptively refines high-frequency details.

Extensive experiments demonstrate state-of-the-art performance in both geometry accuracy and material decomposition. Our contributions are as follows:
\begin{itemize}
    \item A robust geometric reconstruction method for glossy objects that mitigates geometry ambiguities by leveraging geometric priors and split-sum approximation; 
    \item Physically-based material decomposition via the full rendering equation evaluation with Monte Carlo importance sampling;
    \item An adaptive specular compensation mechanism that directionally refines high-frequency details, mitigating fidelity limits in specular rendering.
\end{itemize}
\section{Related Work}
\label{sec:relat}

\textbf{Novel View Synthesis.}
Novel View Synthesis generates unseen scene perspectives from limited input images. Neural Radiance Fields (NeRF)~\cite{mildenhall2021nerf} pioneered photorealistic synthesis through volumetric rendering with implicit neural representations. Subsequent work advanced three directions: geometry enhancement~\cite{wang2021neus}, rendering quality~\cite{barron2021mipnerf,barron2022mipnerf360}, and accelerated training and rendering via hybrid representations such as hash grids and voxels~\cite{muller2022instantngp,chen2022tensorf}.
3D Gaussian Splatting (3DGS)~\cite{kerbl20233dgs} revolutionized NVS with anisotropic 3D Gaussians and tile-based rasterization, achieving state-of-the-art quality and speed. Follow-ups further improved geometric fidelity through surface regularization~\cite{guedon2024sugar, lu2024scaffold}, anti-aliasing~\cite{yu2024mipsplatting}, and texture and gradient optimization~\cite{yu2024gof, chen2024pgsr}. Addressing 3DGS's geometric inconsistencies, 2D Gaussian Splatting (2DGS)~\cite{huang20242dgs} projects disks onto explicit surfaces with local smoothing for view-consistent reconstruction. Our work adopts 2D Gaussian primitives to improve geometric accuracy.

\noindent\textbf{Inverse Rendering of Glossy Objects.}
Inverse rendering estimates scene geometry, materials, and lighting from images but suffers from material-light ambiguities and complex light interactions on glossy surfaces. NeRF-based approaches employ neural radiance fields with ray marching but incur computational bottlenecks from dense sampling. Ref-NeRF~\cite{verbin2022refnerf} simplifies view-dependent effects via directional encoding, yet omits material-light decomposition. TensoIR~\cite{jin2023tensoir} uses tensor factorization for geometry and ray marching for indirect illumination, while ENVIDR~\cite{liang2023envidr} employs surface-based modeling of glossy reflections via a decomposed neural renderer. NeRO~\cite{liu2023nero} reconstructs geometry using split-sum approximation without masks, utilizing integrated directional encoding for illumination, then recovers BRDF and lighting via Monte Carlo sampling. Despite flexibility, these implicit methods incur significant overhead from dense ray sampling.

Recent 3DGS-based methods significantly advance inverse rendering. Pioneering works~\cite{jiang2024gaussianshader, liang2024gsir} apply simplified per-primitive rendering to Gaussians but face geometric and material fidelity limitations. Subsequent methods demonstrate divergent strengths: 3DGS-DR~\cite{ye20243dgsdr} optimizes high-frequency details via deferred shading; Ref-Gaussian~\cite{yao2024refgaussian} uses split-sum approximation for glossy surfaces, inspiring our geometry reconstruction. Parallel developments integrate specialized priors such as microfacet segmentation~\cite{lai2025glossygs}, foundation model supervision~\cite{tong2025gs2dgs}, and diffusion models~\cite{du2024gsid} to reduce geometry ambiguities, motivating our geometry-aware initialization. R3DG~\cite{gao2024r3dg} models the full rendering equation, extended by IRGS~\cite{gu2025irgs} with differentiable 2D Gaussian ray tracing~\cite{Moenne-3DGSRT} for precise visibility and indirect illumination, a technique adopted in our pipeline. Finally, spherical mipmap encoding and directional factorization~\cite{zhang2025refgs, kouros2025rgsdr} enhance specular effects, guiding our specular compensation.
\begin{figure*}
    \vspace{-1em} 
    \centering
    \includegraphics[width=1\linewidth]{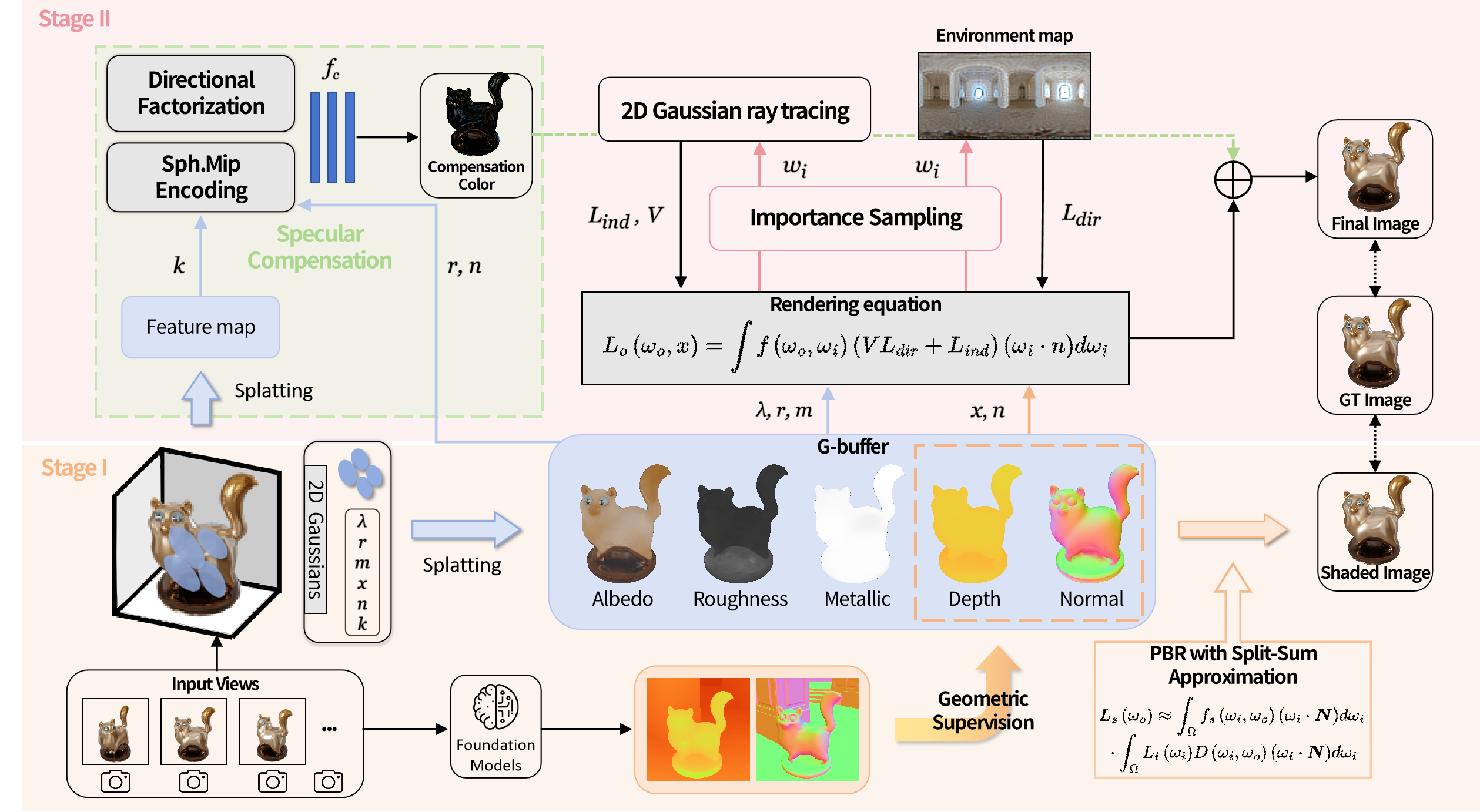}
    \caption{Overview of : Stage I (Sec.~\ref{sec:method-geo}) reconstructs geometry with 2DGS using split-sum shading, supervised by foundation model priors with geometric curvature losses. Stage II (Sec.~\ref{sec:method-ir}) performs inverse rendering on fixed geometry via Monte Carlo sampling, utilizing 2D Gaussian ray tracing for visibility and indirect illumination and spherical mipmap-based compensation for high-frequency specular details. The pipeline's decoupled geometry-material optimization enables high-fidelity relighting.}
    \label{fig:pipline}
    \vspace{-1em} 
\end{figure*}
\vspace{-0.7em}
\section{Preliminary}
\label{sec:prelim}
\subsection{2D Gaussian Splatting}
2D Gaussian Splatting (2DGS)~\cite{huang20242dgs} models scenes using oriented planar disks derived from projected 3D Gaussian distributions, establishing view-consistent geometry with explicit surface normals defined as the direction of steepest density change. Each primitive is parameterized by: a center $\boldsymbol{p} \in \mathbb{R}^3$, orthogonal tangent vectors $\boldsymbol{t}_u$ and $\boldsymbol{t}_v$ defining disk orientation, and scaling factors $s_u$, $s_v$ controlling axial variances. The combined rotation and scaling transformations are represented by a covariance matrix $\boldsymbol{\Sigma}$.

Unlike volumetric rendering, 2DGS employs explicit ray-splat intersections to evaluate Gaussian contributions directly on 2D disks, enabling perspective-correct splatting while mitigating perspective distortion. The Gaussian function is defined as:
\begin{equation}
\label{eq:2dgs}
\mathcal{G}(\boldsymbol{u}) = \exp\left(-\frac{1}{2}(u^2 + v^2)\right),
\end{equation}
where $\boldsymbol{u} = (u, v)$ denotes the local UV coordinates of the ray-disk intersection.

Rendering uses front-to-back alpha blending of depth-sorted Gaussians with degeneracy handling:
\begin{equation}
\label{eq:splatting}
c(\boldsymbol{r}) = \sum_{i=1}^{N} \boldsymbol{c}_i \alpha_i \hat{\mathcal{G}}_i(\boldsymbol{u}) \prod_{j=1}^{i-1} \left(1 - \alpha_j \hat{\mathcal{G}}_j(\boldsymbol{u})\right),
\end{equation}
where $N$ is the number of overlapping Gaussians, $\alpha_i$ the opacity, and $\boldsymbol{c}_i$ the view-dependent color of the $i$-th Gaussian.
\subsection{Physically Based Deferred Shading}
In physically-based Gaussian splatting rendering, shading-rasterization order critically impacts fidelity. We adopt deferred shading to decouple this process: a geometry pass stores attributes in G-buffers, followed by a per-pixel lighting pass. Building on 3DGS-DR's pioneering deferred shading integration~\cite{ye20243dgsdr}, our method extends it to physically-based rendering using Disney BRDF~\cite{burley2012physically}. Each 2D Gaussian carries: albedo $\boldsymbol{\lambda} \in [0,1]^3$; metallic $m \in [0,1]$; roughness $r \in [0,1]$; surface normal $\boldsymbol{n} = \boldsymbol{t_u} \times \boldsymbol{t_v}$; position $\boldsymbol{x}_i$; and feature vector $k_i$ that aggregates into the feature map $K$ utilized in the second stage (Sec.~\ref{sec:method-ir}) to enhance specular effects.

We render attributes into G-buffers using alpha-blending from Eq.~\ref{eq:splatting}, replacing color $\boldsymbol{c}_i$ with $\boldsymbol{f}_i$:
\begin{equation}
\label{eq:splatting-feas}
\boldsymbol{F}=\sum_{i=1}^N \boldsymbol{f}_i \alpha_i \hat{\mathcal{G}}_i(\boldsymbol{u}(\boldsymbol{r})) \prod_{j=1}^{i-1}\left(1-\alpha_j \hat{\mathcal{G}}_j(\boldsymbol{u}(\boldsymbol{r}))\right),
\end{equation}
where $\boldsymbol{f}_i=\left[\boldsymbol{\lambda}_i, m_i, r_i, \boldsymbol{n}_i, \boldsymbol{x}_i, k_i\right]^{\top}$ contains per-Gaussian properties, yielding aggregated attributes $\boldsymbol{F}=$ $[\boldsymbol{\Lambda}, M, R, \boldsymbol{N}, \boldsymbol{D}, K]^{\top}$. 
Using aggregated G-buffer attributes, we compute outgoing radiance $L_o$ at $\boldsymbol{x}$ with normal $\boldsymbol{n}$ via the rendering equation~\cite{Kajiya_1998}:
\begin{equation}
L_o\left(\boldsymbol{\omega}_o, \boldsymbol{x}\right) = \int_{\Omega} f\left(\boldsymbol{\omega}_o, \boldsymbol{\omega}_i, \boldsymbol{x}\right) L_{\mathrm{i}}\left(\boldsymbol{\omega}_i, \boldsymbol{x}\right) (\boldsymbol{\omega}_i \cdot \boldsymbol{n})  d \boldsymbol{\omega}_i,
\label{eq:renderingEquation}
\end{equation}
where $f$ is the Disney BRDF, consisting of a diffuse term $f_d$
and a specular term $f_s$:
\begin{equation}
f_d = \frac{\boldsymbol{\Lambda}}{\pi}(1 - M)(1 - F_d),
\end{equation}
\begin{equation}
f_s\left(\boldsymbol{\omega}_i, \boldsymbol{\omega}_o, \boldsymbol{x}\right) = \frac{DFG}{4\left(\boldsymbol{\omega}_i \cdot \boldsymbol{n}\right)\left(\boldsymbol{\omega}_o \cdot \boldsymbol{n}\right)}.
\label{eq:disney_spec}
\end{equation}
Here, $D$, $F$, and $G$ represent the microsurface normal distribution, Fresnel reflectance, and geometry attenuation terms, respectively, with the diffuse Fresnel factor $F_d$ enforcing energy conservation by deducting specular-reflected energy. These terms are computed using the stored roughness $R$ and metallic $M$ values, following the energy-conserving properties of the Disney BRDF model~\cite{burley2012physically}.
\section{Method}
\label{sec:method}
We present a novel inverse rendering framework for reconstructing reflective objects from multi-view RGB images under unknown illumination. As illustrated in Fig.~\ref{fig:pipline}, our approach decomposes the problem into two sequential stages: First, geometry reconstruction leverages deferred Gaussian splatting with split-sum approximation for efficient physically-based rendering while incorporating geometric priors from foundation models to mitigate geometry ambiguities (Sec.~\ref{sec:method-geo}). Second, with object geometry fixed, we refine material properties through Monte Carlo importance sampling of the full rendering equation to optimize BRDF parameters, enhanced by a spherical mipmap-based specular compensation mechanism for complex specular effects (Sec.~\ref{sec:method-ir}).
\subsection{Stage I: Geometry reconstruction}
\label{sec:method-geo}
\noindent\textbf{Split-sum Approximation.}
Within deferred shading using aggregated per-pixel material attributes, we compute the specular term via split-sum approximation~\cite{munkberg2022extracting} to mitigate its computational cost. This decomposes the rendering integral into:
\begin{equation}
\label{eq:split_sum}
L_s(\boldsymbol{\omega}_o) \approx \underbrace{\int_{\Omega} f_s(\cdot) (\boldsymbol{\omega}_i\!\cdot\!\boldsymbol{n}) d\boldsymbol{\omega}_i}_{\text{BRDF factor}} \cdot \underbrace{\int_{\Omega} L_i(\cdot) D(\cdot)(\boldsymbol{\omega}_i\!\cdot\!\boldsymbol{n})  d\boldsymbol{\omega}_i}_{\text{Lighting factor}}.
\end{equation}
The BRDF factor, dependent on material roughness $R$ and normal angle $(\boldsymbol{\omega}_i \cdot \boldsymbol{n})$, is precomputed into a 2D lookup texture. The lighting factor integrates environmental radiance over the specular lobe using trilinear interpolation in pre-filtered Mipmap cubemaps, indexed by reflected direction $\boldsymbol{\omega}_r = 2(\boldsymbol{n} \cdot \boldsymbol{\omega}_o)\boldsymbol{n} - \boldsymbol{\omega}_o$ and roughness $R$. 
While computationally efficient, this approximation inherently limits high-frequency specular modeling, particularly on low-roughness surfaces, necessitating refinement with the full rendering equation.

\noindent\textbf{Supervision from Foundation Models.}
To mitigate geometry ambiguities in specular surface reconstruction, we leverage robust geometric priors from large-scale vision foundation models~\cite{Ke_Obukhov_Huang_Metzger_Daudt_Schindler_2023, garcia2025fine}. These models generate monocular depth $\tilde{D}$ and normal $\tilde{N}$ estimates invariant to view-dependent effects, providing reliable supervision even for highly reflective surfaces where multi-view consistency fails.

\begin{figure}[h!]
    \vspace{-1em}
    \centering
    \includegraphics[width=1\linewidth]{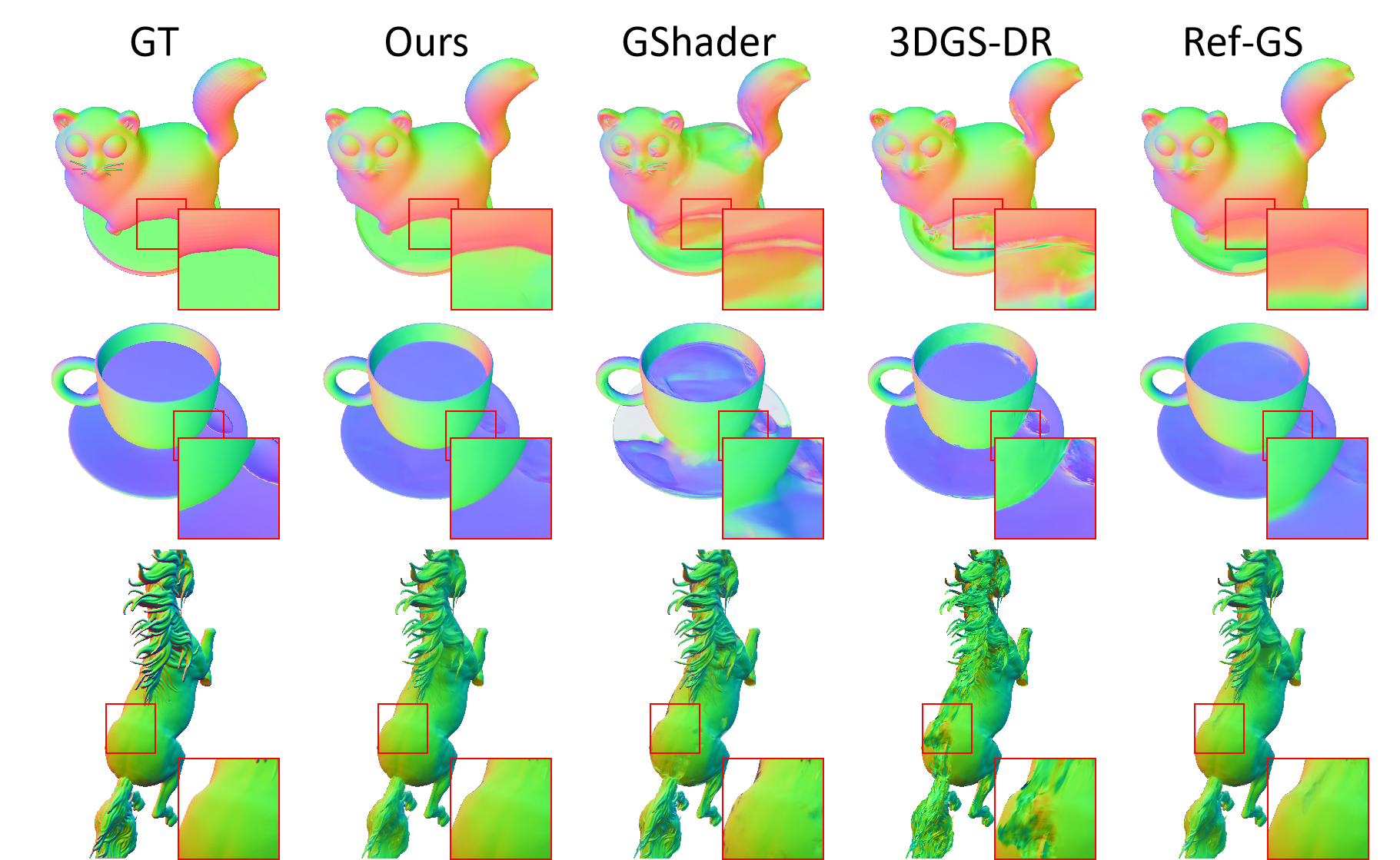}
    \caption{Per-scene qualitative comparisons of normals.}
    \label{fig:normal_compa}
    \vspace{-1em}
\end{figure}

We incorporate these priors as regularization during Gaussian optimization. The predicted normals $\tilde{N}$ supervise rendered normals $\hat{N}$ via a dual loss:
\begin{equation}
\label{eq:L-geo-n}
\mathcal{L}_{\text{geo-n}} = \|\hat{N} - \tilde{N}\|_1 + \lambda \left(1 - \frac{\hat{N} \cdot \tilde{N}}{\|\hat{N}\| \|\tilde{N}\|}\right),
\end{equation}
where $\lambda$ balances magnitude and angular alignment. For depth supervision, we employ a scale-invariant formulation~\cite{ranftl2020towards} with predicted normals $\tilde{D}$ and rendered normals $\hat{D}$:
\begin{equation}
\label{eq:L-geo-d}
\mathcal{L}_{\text{geo-d}} = \min_{\omega, b} \sum_{p} \left[ (\omega \hat{D} + b) - \tilde{D} \right]^2,
\end{equation}
with $\omega$ and $b$ optimized per-view via least squares. As demonstrated in Fig.~\ref{fig:normal_compa}, this geometric regularization explicitly enhances curvature continuity while suppressing surface noise from specular interference.

\noindent\textbf{Training Loss.}
The complete training objective integrates physical rendering losses with our geometric priors:
\begin{equation}
\begin{split}
\mathcal{L}^1 = & \mathcal{L}_{\mathrm{c}} 
+ \lambda_{\mathrm{n}} \mathcal{L}_{\mathrm{n}} 
+ \lambda_{\mathrm{o}} \mathcal{L}_{\mathrm{o}} 
+ \lambda_{\text{smooth}} \mathcal{L}_{\text{smooth}} \\
& + \lambda_{\text{geo-n}} \mathcal{L}_{\text{geo-n}} 
+ \lambda_{\text{geo-d}} \mathcal{L}_{\text{geo-d}}.
\end{split}
\end{equation}
Following 2DGS~\cite{huang20242dgs}, we utilize its core components including the RGB reconstruction loss $\mathcal{L}_\mathrm{c} = \|c_{\text{render}} - C_{gt}\|_1$ and the normal alignment loss defined as $\mathcal{L}_\mathrm{n} = 1 - \tilde{\boldsymbol{N}}^{\mathrm{T}} \boldsymbol{N}$.
$\mathcal{L}_\mathrm{o}$ is a binary cross-entropy loss constraining the geometry using the provided object mask $\mathcal{M}$ defined as $\mathcal{L}_{\mathrm{o}}=-\mathcal{M} \log \mathcal{O}-(1-\mathcal{M}) \log (1-\mathcal{O})$, where $\mathcal{O}=\sum_{i=1}^N T_i \alpha_i$ denotes the accumulated opacity map. 
To provide joint regularization for both normals and depth, we incorporate an edge-aware smoothness term 
$\mathcal{L}_\text{smooth} = \|\nabla \boldsymbol{N}\| \exp\left(-\|\nabla C_{gt}\|\right)$.
Finally, our novel geometric regularizers $\mathcal{L}_\text{geo-n}$ and $\mathcal{L}_\text{geo-d}$, 
defined in Eq.~\ref{eq:L-geo-n} and Eq.~\ref{eq:L-geo-d}, enforce alignment with foundation model priors.

\begin{table*}[h!]
\vspace{-1em}
\centering
\captionsetup{width=0.9\textwidth}
\caption{Quantitative comparison of average novel view synthesis metrics on synthesized test views.
Deeper red indicates better performance.
Ours(geo) corresponds to results from our first-stage geometric reconstruction (Sec.~\ref{sec:method-geo}), while Ours(ir) corresponds to those from our second-stage inverse rendering (Sec.~\ref{sec:method-ir}).}
\label{tab:quality_compa}
\resizebox{0.95\textwidth}{!}{%
\begin{tabular}{l|ccc|ccc|ccc}
\hline
\hline
\textbf{Datasets} & \multicolumn{3}{c|}{\textbf{Shiny Synthetic\cite{verbin2022refnerf}}}                                                 & \multicolumn{3}{c|}{\textbf{Glossy Synthetic\cite{liu2023nero}}}                                                & \multicolumn{3}{c}{\textbf{Shiny Real\cite{verbin2022refnerf}}}                                                       \\ \hline
                  & PSNR ↑                        & SSIM ↑                        & LPIPS ↓                       & PSNR ↑                        & SSIM ↑                        & LPIPS ↓                       & PSNR ↑                        & SSIM ↑                        & LPIPS ↓                       \\ \hline
Ref-NeRF\cite{verbin2022refnerf}          & 33.13                         & 0.961                         & 0.080                         & 25.65                         & 0.905                         & 0.112                         & 23.62                         & 0.646                         & 0.239                         \\
ENVIDR\cite{liang2023envidr}            & 33.46                         & \cellcolor[HTML]{FD6864}0.979 & \cellcolor[HTML]{FD6864}0.046 & 29.06                         & 0.947                         & 0.060                         & 23.00                         & 0.606                         & 0.332                         \\
2DGS\cite{huang20242dgs}              & 29.58                         & 0.946                         & 0.084                         & 26.07                         & 0.918                         & 0.088                         & 24.15                         & 0.661                         & 0.292                         \\
GShader\cite{jiang2024gaussianshader}           & 31.97                         & 0.958                         & 0.067                         & 27.11                         & 0.922                         & 0.145                         & 23.46                         & 0.647                         & 0.257                         \\
R3DG\cite{gao2024r3dg}              & 27.77                         & 0.926                         & 0.112                         & 24.13                         & 0.892                         & 0.106                         & 21.98                         & 0.619                         & 0.349                         \\
3DGS-DR\cite{ye20243dgsdr}      & 34.00                         & 0.972                         & 0.059                         & 28.73                         & 0.948                         & 0.058                         & 24.37                         & 0.678                         & \cellcolor[HTML]{FFCCC9}0.232 \\
Ref-Gaussian\cite{yao2024refgaussian}            & \cellcolor[HTML]{FFCCC9}34.66 & 0.972                         & 0.055                         & \cellcolor[HTML]{FFCCC9}30.77 & \cellcolor[HTML]{FFCCC9}0.962 & \cellcolor[HTML]{FD6864}0.045 & \cellcolor[HTML]{FFCCC9}24.71 & \cellcolor[HTML]{FFCCC9}0.691 & 0.263                         \\
IRGS\cite{gu2025irgs}              & 28.39                         & 0.932                         & 0.110                         & 24.40                         & 0.892                         & 0.109                         & 21.38                         & 0.425                         & 0.486                         \\
Ours(geo)          & \cellcolor[HTML]{FD6864}35.03 & \cellcolor[HTML]{FFCCC9}0.975 & \cellcolor[HTML]{FFCCC9}0.055 & \cellcolor[HTML]{FD6864}30.83 & \cellcolor[HTML]{FD6864}0.962 & \cellcolor[HTML]{FFCCC9}0.048 & \cellcolor[HTML]{FD6864}26.56 & \cellcolor[HTML]{FD6864}0.783 & \cellcolor[HTML]{FD6864}0.178 \\
Ours(ir)          & 32.21                         & 0.959                         & 0.078                         & 29.38                         & 0.947                         & 0.060                         & 24.47                         & 0.687                         & 0.299                         \\ \hline
\hline
\end{tabular}%
}
\end{table*}

\subsection{Stage II: Inverse Rendering}
\label{sec:method-ir}
Building upon the reconstructed geometry from Stage I~\ref{sec:method-geo}, we now refine the initial BRDF estimation through physically accurate inverse rendering. This stage employs the full rendering equation to precisely optimize material parameters, including metalness $m$, albedo $a$, and roughness $\rho$, while leveraging Monte Carlo importance sampling for efficient incident radiance computation and rendering equation evaluation.

\noindent\textbf{Sampling Strategy.} 
We implement two physically-based importance sampling strategies for optimal variance reduction. For the diffuse component, we follow the Lambertian distribution with sampling probability:
\begin{equation}
p_d(\omega_i) = \frac{\cos\theta}{\pi},
\end{equation}
while for the specular component, we employ GGX normal distribution-based sampling~\cite{cook1982reflectance}:
\begin{equation}
p_s(\omega_i) = \frac{D(\boldsymbol{h}) \cos\theta}{4(\boldsymbol{\omega_o} \cdot \boldsymbol{h})}.
\end{equation}
To minimize variance in transitional regions between diffuse and specular responses, we integrate both sampling strategies through a balance heuristic~\cite{veach1998robust}. Defining strategy proportions $\pi_d = N_d / (N_d + N_s)$ and $\pi_s = N_s / (N_d + N_s)$, the weight for a sample $\omega_i$ from strategy $k \in \{d, s\}$ is computed as:
\begin{equation}
w_k(\omega_i) = \frac{\pi_k p_k(\omega_i)}{\sum_{m \in \{d,s\}} \pi_m p_m(\omega_i)}.
\end{equation}
The final radiance $L_o^{\mathrm{phys}}$ estimate combines both contributions:
\begin{align}
L_o^{\mathrm{diffuse}} &= \frac{1}{N_d} \sum_{i=1}^{N_d} \frac{f_d L_i (\boldsymbol{n} \cdot \omega_i) w_d}{p_d}, \\
L_o^{\mathrm{specular}} &= \frac{1}{N_s} \sum_{j=1}^{N_s} \frac{f_s L_j (\boldsymbol{n} \cdot \omega_j) w_s}{p_s}, \\
L_o^{\mathrm{pbr}} &= L_o^{\mathrm{diffuse}} + L_o^{\mathrm{specular}}.
\end{align}
This formulation adaptively balances sample contributions based on local material properties, significantly reducing noise while maintaining physical accuracy.

{
\setlength{\parindent}{0pt}
\textbf{Light parametrization.} The incident radiance $L_{\mathrm{i}}(\boldsymbol{\omega}_i, \boldsymbol{x})$ at surface point $\boldsymbol{x}$ along direction $\boldsymbol{\omega}_i$ (Eq.~\ref{eq:renderingEquation}) decomposes into direct radiance from distant sources and indirect radiance from scene surfaces:
\begin{equation}
L_{\mathrm{i}}\left(\boldsymbol{\omega}_i, \boldsymbol{x}\right)=V\left(\boldsymbol{\omega}_i, \boldsymbol{x}\right) L_{\mathrm{dir}}\left(\boldsymbol{\omega}_i\right)+L_{\mathrm{ind}}\left(\boldsymbol{\omega}_i, \boldsymbol{x}\right), 
\end{equation}
where direct radiance $L_{\mathrm{dir}}$ is parameterized via an environment cubemap, while visibility $V$ and indirect radiance $L_{\mathrm{ind}}$ are computed through IRGS's differentiable 2D Gaussian ray tracing~\cite{gu2025irgs}.
}
\begin{figure*}[h!]
    \vspace{-1em}
    \centering
    \includegraphics[width=0.9\linewidth]{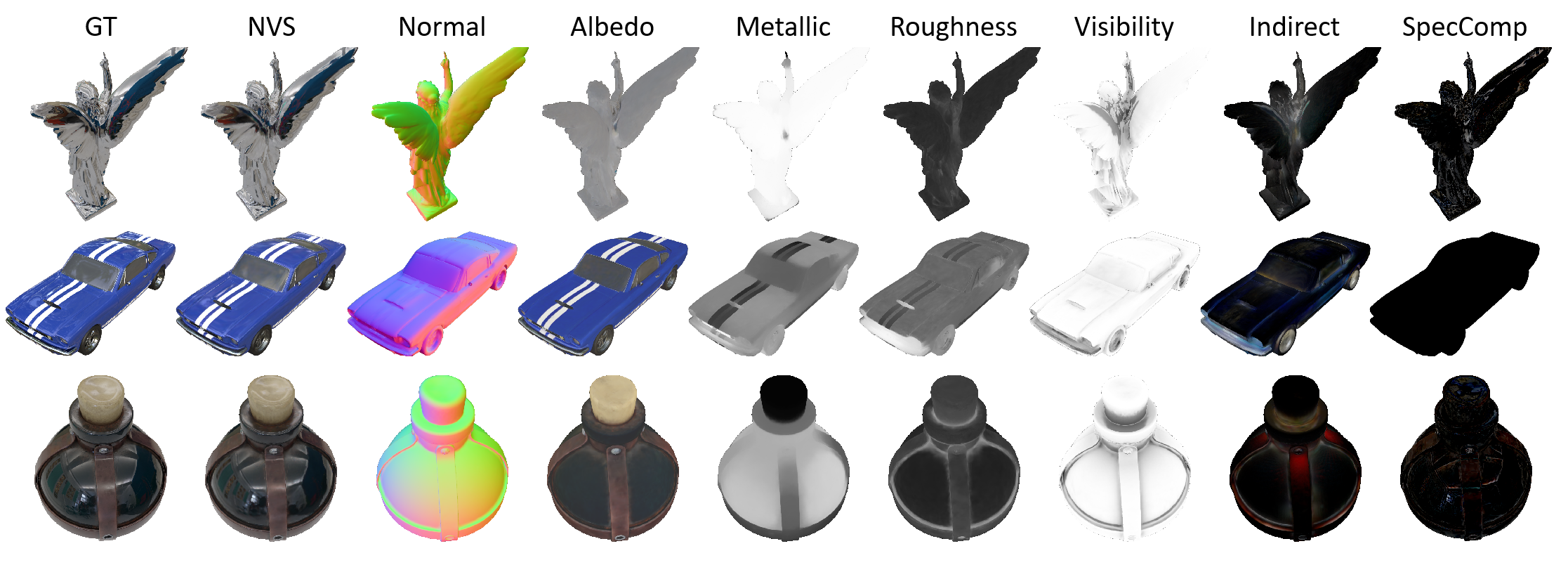}
    \captionsetup{width=0.9\linewidth}
    \caption{Qualitative decomposition results of our model.}
    \label{fig:results}
    \vspace{-1em}
\end{figure*}

\begin{figure}
    \centering
    \includegraphics[width=1\linewidth]{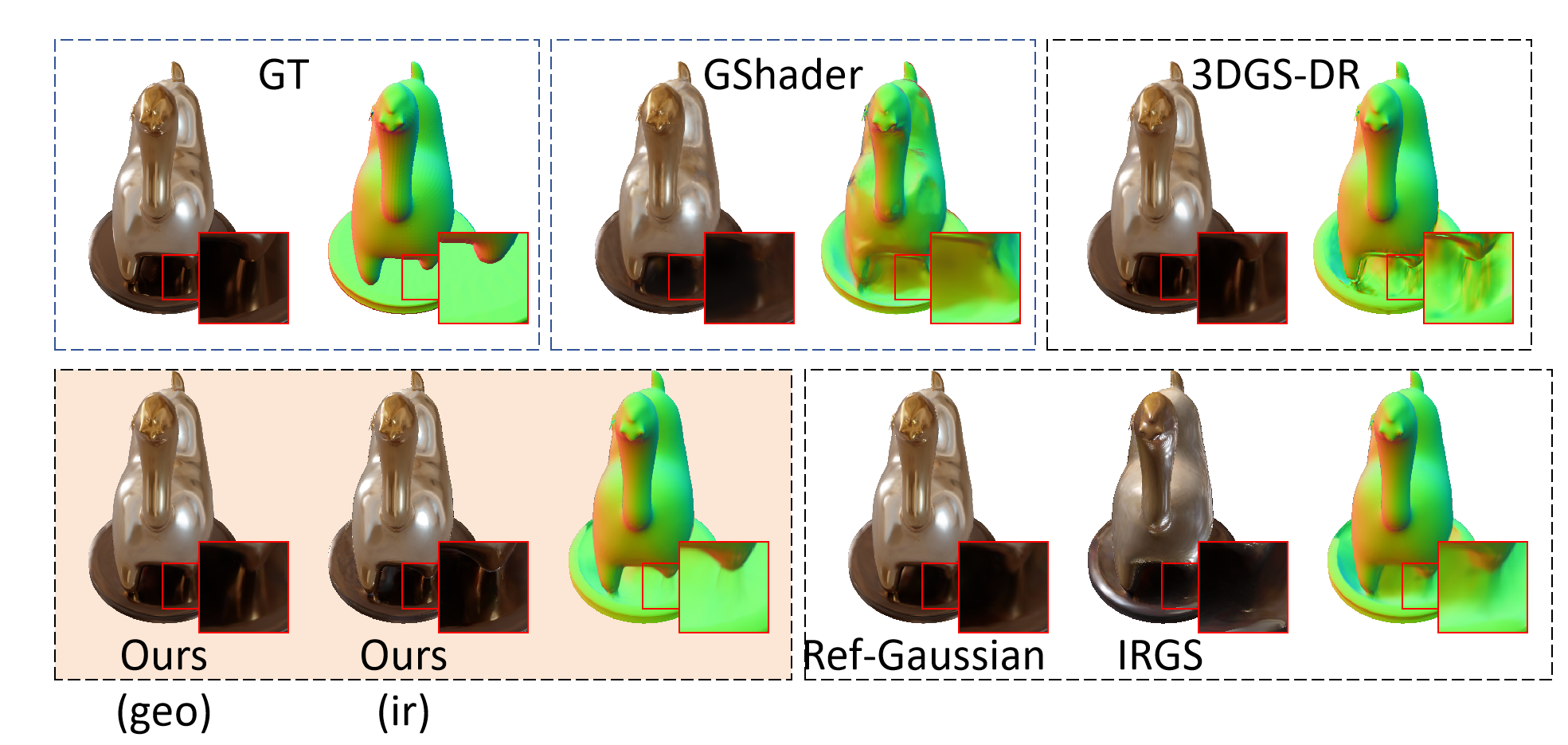}
    \captionsetup{width=0.9\linewidth}
    \caption{Qualitative comparisons of NVS renderings and normal maps, focusing on geometric accuracy and indirect illumination in inter-reflection regions.}
    \label{fig:nvs_normal_compa}
    \vspace{-1em}
\end{figure}

\noindent\textbf{Specular Compensation.}
To mitigate high-frequency artifacts from Monte Carlo sampling variance, specifically material parameter noise and specular-diffuse separation errors, we introduce a compensation mechanism. Drawing inspiration from the directional factorization paradigm~\cite{zhang2025refgs}, 
we leverage the blended feature map $K$ stored in the G-buffer (Eq.~\ref{eq:splatting-feas}) and a directionally encoded feature $\boldsymbol{h}$ queried via spherical Mip-grid $\mathcal{M}$:
\begin{equation}
\boldsymbol{h} = \mathcal{M}\left( \theta_r(\boldsymbol{N}, \boldsymbol{\omega}_o), \phi_r(\boldsymbol{N}, \boldsymbol{\omega}_o), R \right),
\end{equation}
where $(\theta_r, \phi_r)$ are spherical coordinates of $\boldsymbol{\omega}_r$, computed using the blended normal $\boldsymbol{N}$ from the G-buffer and viewing direction $\boldsymbol{\omega}_o$, with $R$ representing the optimized surface roughness.
The compensation radiance $\boldsymbol{L}_c$ is synthesized by:
\begin{equation}
\boldsymbol{L}_c = f_c \left( K, \boldsymbol{h} \right).
\end{equation}
This term refines physical shading through additive blending:
\begin{equation}
L_o^{\mathrm{final}} = L_o^{\mathrm{pbr}} + \boldsymbol{L}_c.
\end{equation}
It is noteworthy that $\boldsymbol{L}_c$ is explicitly disabled during relighting with novel illumination and operates exclusively within the inverse rendering optimization loop to enhance reconstruction fidelity for difficult-to-model specular phenomena.

\noindent\textbf{Training Loss.}
We define the total loss $\mathcal{L}^2$ as:
\begin{equation}
\mathcal{L}^2 = \mathcal{L}_{\mathrm{c}} + \lambda_{\text{smooth}} \mathcal{L}_{\text{smooth}} + \lambda_{\text{light}} \mathcal{L}_{\text{light}}.
\end{equation}
We retaining the RGB reconstruction loss $\mathcal{L}_{\mathrm{c}}$ and smoothness regularization $\mathcal{L}_{\text{smooth}}$ from the first stage. The edge-aware smoothness regularization $\mathcal{L}_{\text{smooth}}$ is now notably applied to rendered material properties including albedo ($\boldsymbol{\Lambda}$), roughness ($R$), and metallic ($M$) maps to enforce coherent surface characteristics. The lighting regularization term $\mathcal{L}_{\text{light}}$ imposes a neutral white prior on diffuse illumination $L_{\text{diffuse}} = \frac{1}{N_{\mathrm{r}}} \sum_{i=1}^{N_{\mathrm{r}}} L\left(\boldsymbol{\omega}_i, \boldsymbol{x}\right)$ by minimizing RGB deviations from their mean value, formulated as $\mathcal{L}_{\text{light}} = \sum_c \left( L_{\text{diffuse}}^c - \frac{1}{3} \sum_{c'} L_{\text{diffuse}}^{c'} \right)$ where $c \in \{R, G, B\}$. Following IRGS~\cite{gu2025irgs} for computational efficiency, we restrict rendering equation evaluations to a subset of pixels per iteration, sampling $\left\lfloor N_{\text{rays}} / N_{\mathrm{r}} \right\rfloor$ pixels through a maximum ray budget $N_{\text{rays}}$ to enable high-quality estimation with large $N_{\mathrm{r}}$.

Our physically-based inverse rendering framework achieves precise material decomposition through joint BRDF-geometry optimization. The recovered materials maintain intrinsic physical consistency, enabling faithful relighting and high-fidelity rendering under novel illumination.
\section{Experiment}
\label{sec:exp}
\noindent\textbf{Datasets and metrics.}
We utilize two synthetic datasets, Shiny Blender~\cite{verbin2022refnerf} and Glossy Synthetic~\cite{liu2023nero}, along with the real-world Shiny Real dataset~\cite{verbin2022refnerf} to evaluating novel view synthesis of reflective objects. For quantitative assessment, we employ three standard image quality metrics: PSNR, SSIM~\cite{ssim}, and LPIPS ~\cite{lpips}, complemented by Mean Angular Error (MAE) for normal estimation accuracy.

\begin{table}[]
\centering
\captionsetup{width=0.99\linewidth}
\caption{Quantitative comparison on normal maps}
\label{tab:normal_compa}
\resizebox{0.99\columnwidth}{!}{%
\begin{tabular}{l|cccc}
\hline
\hline
Metrics & \multicolumn{4}{c}{Model}                                  \\ \hline
        & GShader\cite{jiang2024gaussianshader} & 3DGS-DR\cite{ye20243dgsdr} & Ref-Gaussian\cite{yao2024refgaussian} & Ours                          \\ \hline
MAE ↓   & 4.74    & 9.49    & 2.28   & \cellcolor[HTML]{FD6864}2.16  \\
SSIM ↑  & 0.853   & 0.824   & 0.924  & \cellcolor[HTML]{FD6864}0.924 \\
LPIPS ↓ & 0.108   & 0.172   & 0.073  & \cellcolor[HTML]{FD6864}0.072 \\ \hline
\hline
\end{tabular}%
}
\vspace{-1em}
\end{table}

\begin{table*}[h!]
\vspace{-1em}
\centering
\captionsetup{width=0.9\textwidth}
\caption{Quantitative comparisons of relighting results in terms of PSNR↑ on the Glossy Synthetic dataset~\cite{liu2023nero}.}
\label{tab:relight_psnr}
\renewcommand{\arraystretch}{0.9}
\footnotesize
\resizebox{0.85\textwidth}{!}{%
\begin{tabular}{lccccccccc}
\hline
\hline
\multicolumn{1}{l|}{\textbf{Datasets}} & \textbf{angel} & \textbf{bell} & \textbf{cat} & \textbf{horse} & \textbf{luyu} & \textbf{potion} & \textbf{tbell} & \textbf{teapot} & \textbf{avg.} \\ \hline
\multicolumn{10}{c}{\textbf{corridor}} \\ \hline
\multicolumn{1}{l|}{GShader\cite{jiang2024gaussianshader}} & 21.83 & 22.98 & 16.42 & \cellcolor[HTML]{FD6864}26.42 & 16.74 & 14.99 & 18.74 & 20.35 & 19.81 \\
\multicolumn{1}{l|}{Ref-Gaussian\cite{yao2024refgaussian}} & 20.97 & 23.02 & 20.57 & 25.45 & 20.34 & 20.22 & 21.51 & 22.59 & 21.83 \\
\multicolumn{1}{l|}{IRGS\cite{gu2025irgs}} & 20.31 & 21.16 & 22.40 & 22.25 & 24.65 & 24.53 & 20.71 & 20.05 & 22.01 \\
\multicolumn{1}{l|}{Ours} & \cellcolor[HTML]{FD6864}23.92 & \cellcolor[HTML]{FD6864}25.76 & \cellcolor[HTML]{FD6864}27.32 & 25.80 & \cellcolor[HTML]{FD6864}25.11 & \cellcolor[HTML]{FD6864}27.42 & \cellcolor[HTML]{FD6864}24.61 & \cellcolor[HTML]{FD6864}26.05 & \cellcolor[HTML]{FD6864}25.75 \\ \hline
\multicolumn{10}{c}{\textbf{golf}} \\ \hline
\multicolumn{1}{l|}{GShader\cite{jiang2024gaussianshader}} & 21.04 & 21.40 & 14.84 & 25.01 & 14.85 & 12.65 & 16.80 & 18.50 & 18.14 \\
\multicolumn{1}{l|}{Ref-Gaussian\cite{yao2024refgaussian}} & 22.32 & 23.05 & 21.57 & 25.87 & 19.41 & 19.85 & 19.96 & 21.54 & 21.69 \\
\multicolumn{1}{l|}{IRGS\cite{gu2025irgs}} & 20.58 & 20.89 & 21.96 & 22.24 & 22.64 & 22.52 & 18.90 & 18.56 & 21.04 \\
\multicolumn{1}{l|}{Ours} & \cellcolor[HTML]{FD6864}26.37 & \cellcolor[HTML]{FD6864}26.99 & \cellcolor[HTML]{FD6864}27.35 & \cellcolor[HTML]{FD6864}26.28 & \cellcolor[HTML]{FD6864}24.85 & \cellcolor[HTML]{FD6864}26.14 & \cellcolor[HTML]{FD6864}23.91 & \cellcolor[HTML]{FD6864}26.19 & \cellcolor[HTML]{FD6864}26.01 \\ \hline
\multicolumn{10}{c}{\textbf{neon}} \\ \hline
\multicolumn{1}{l|}{GShader\cite{jiang2024gaussianshader}} & 21.16 & 21.41 & 16.74 & 24.12 & 16.87 & 15.59 & 18.94 & 19.07 & 19.24 \\
\multicolumn{1}{l|}{Ref-Gaussian\cite{yao2024refgaussian}} & 21.20 & \cellcolor[HTML]{FD6864}21.85 & 20.56 & 23.32 & 20.41 & 20.23 & 21.65 & 21.64 & 21.36 \\
\multicolumn{1}{l|}{IRGS\cite{gu2025irgs}} & 20.30 & 19.75 & 21.33 & 21.29 & 20.27 & 20.69 & 18.71 & 18.64 & 20.12 \\
\multicolumn{1}{l|}{Ours} & \cellcolor[HTML]{FD6864}21.72 & 21.36 & \cellcolor[HTML]{FD6864}25.31 & \cellcolor[HTML]{FD6864}24.52 & \cellcolor[HTML]{FD6864}21.65 & \cellcolor[HTML]{FD6864}23.66 & \cellcolor[HTML]{FD6864}21.87 & \cellcolor[HTML]{FD6864}22.93 & \cellcolor[HTML]{FD6864}22.88 \\ \hline
\hline
\end{tabular}%
}
\vspace{-1em}
\end{table*}

\noindent\textbf{Implementation details.}
All experiments run on a single NVIDIA RTX 3090 GPU. Our training pipeline follows the two-stage process in Sec~\ref{sec:method}. The first stage trains for 50,000 iterations with hyperparameters: $\lambda_{\mathrm{n}} = 0.05$, $\lambda_{\mathrm{d}} = 0.05$, $\lambda_{\text{smooth}} = 0.01$, $\lambda_{\text{geo-n}} = 0.005$, and $\lambda_{\text{geo-d}} = 0.005$. We use Marigold fine-tuned for normal and depth estimation due to its superior object-level prediction. The environmental lighting employs a $6 \times 128 \times 128$ mipmap cubemap with 3 RGB channels. The second stage performs 20,000 iterations of inverse rendering with $\lambda_{\text{smooth}} = 2.0$ and $\lambda_{\text{light}} = 0.01$. The learning rates are set to 0.0075 for albedo, 0.005 for roughness, 0.005 for metallic, and 0.01 for cubemap. Ray tracing uses IRGS~\cite{gu2025irgs} hyperparameters while other settings follow 2DGS~\cite{huang20242dgs}. The output at this stage enables direct relighting with ray samples increased to 2048 for improved quality. For enhanced novel view synthesis, we perform up to 80,000 specular compensation iterations with geometry, materials and lighting frozen. Only the spherical encoding components are updated, including an $8 \times 512 \times 512$ mipmap with 16 feature channels and a shallow MLP that has two 256-unit hidden layers. The total training without specular compensation takes approximately 80 minutes: 40 minutes for Stage I~\ref{sec:method-geo} and 40 minutes for Stage II~\ref{sec:method-ir}. Specular compensation requires an additional hour.

\begin{figure}[]
    \centering
    \includegraphics[width=1\linewidth]{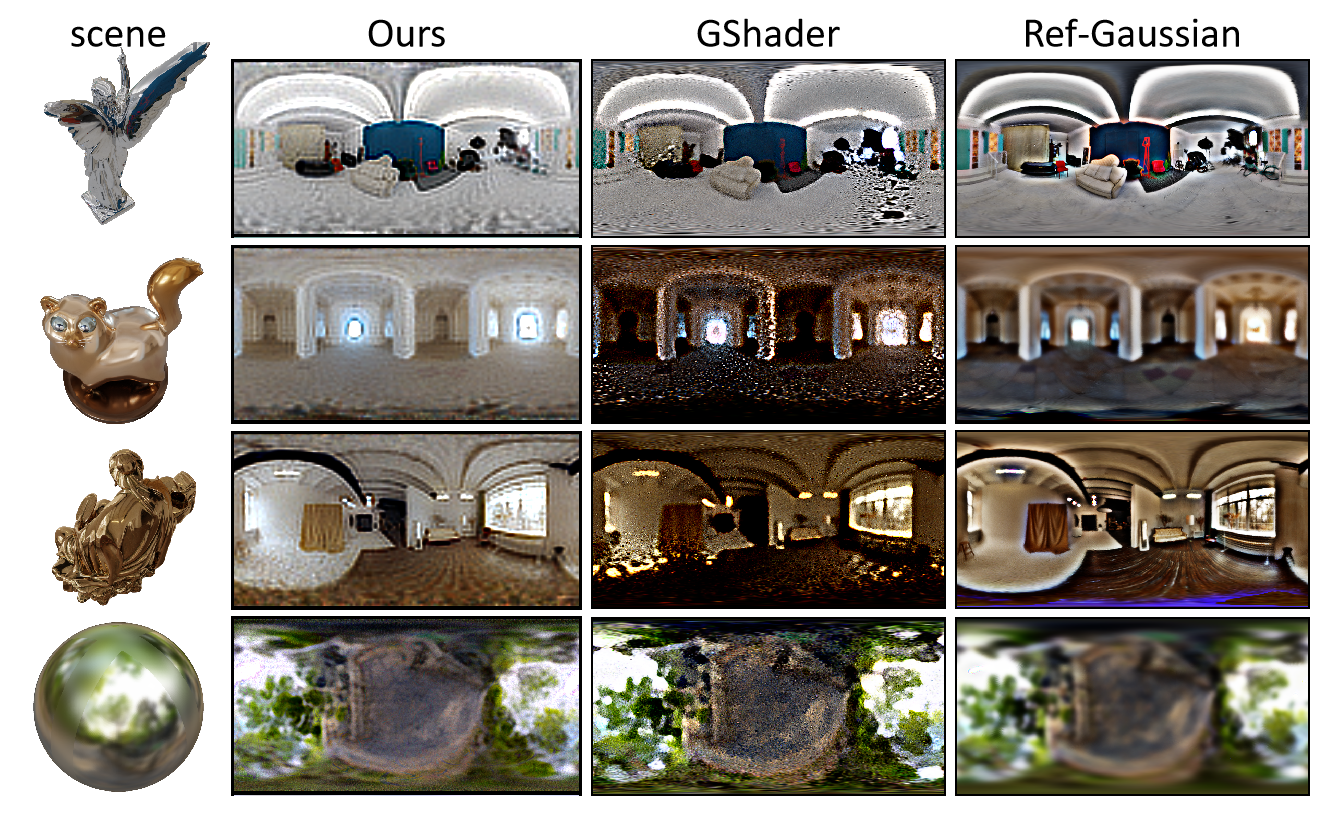}
    \captionsetup{width=0.9\linewidth}
    \caption{Qualitative comparisons of the estimated environment maps.}
    \label{fig:envmap_compa}
    \vspace{-1em}
\end{figure}
\subsection{Comparison}
\textbf{Novel View Synthesis.} 
Tab.~\ref{tab:quality_compa} shows quantitative comparisons against SOTA across multiple datasets, including average metric scores per dataset. Our first-stage (Sec.~\ref{sec:method-geo}) consistently outperforms SOTA across all metrics, with notable gains on Glossy Synthetic~\cite{liu2023nero} due to geometric priors. Although second-stage results (Sec.~\ref{sec:method-ir}) exhibit a modest decrease in NVS performance, they yield significantly more accurate material decomposition, as validated by subsequent relighting experiments. 
Fig.~\ref{fig:results} illustrates the output decomposition: geometry (via normals), material properties, visibility, indirect illumination, and specular compensation. Our normals are smoother and more accurate, while material decomposition adheres better to physical constraints, effectively disentangling geometry and materials for photorealistic rendering.
Fig.~\ref{fig:envmap_compa} compares the estimated environment maps. Our method effectively disentangles illumination from scene properties. Versus Ref-Gaussian~\cite{yao2024refgaussian} and GShader~\cite{jiang2024gaussianshader}, we obtain more refined maps with sharper details, fewer artifacts, and closer to ground truth.

{
\setlength{\parindent}{0pt} 
\textbf{Geometry Reconstruction.}  
Tab.~\ref{tab:normal_compa} quantitatively validates our superior normal map quality, attributable to first-stage geometry optimization. 
Fig.~\ref{fig:normal_compa} and Fig.~\ref{fig:nvs_normal_compa} further illustrate our advantage. The qualitative comparison in Fig.~\ref{fig:normal_compa} demonstrates our method's comprehensive detail capture, such as the base of a cat, cup bottom, and horse back.
Fig.~\ref{fig:nvs_normal_compa} compares our approach to competitors on glossy objects. While competitors achieve adequate novel view synthesis, they fail to reconstruct accurate geometry or precise indirect illumination in inter-reflection regions. In contrast, our method leverages geometric priors to recover precise inter-reflection geometry and generates physically consistent indirect illumination via ray tracing, producing realistic light bounces (red boxes) absent in competitors. Our solution consistently delivers accurate normals and photorealistic indirect illumination, critical for high-quality novel view synthesis in reflective scenes.
}

\begin{figure}
    \centering
    \includegraphics[width=1\linewidth]{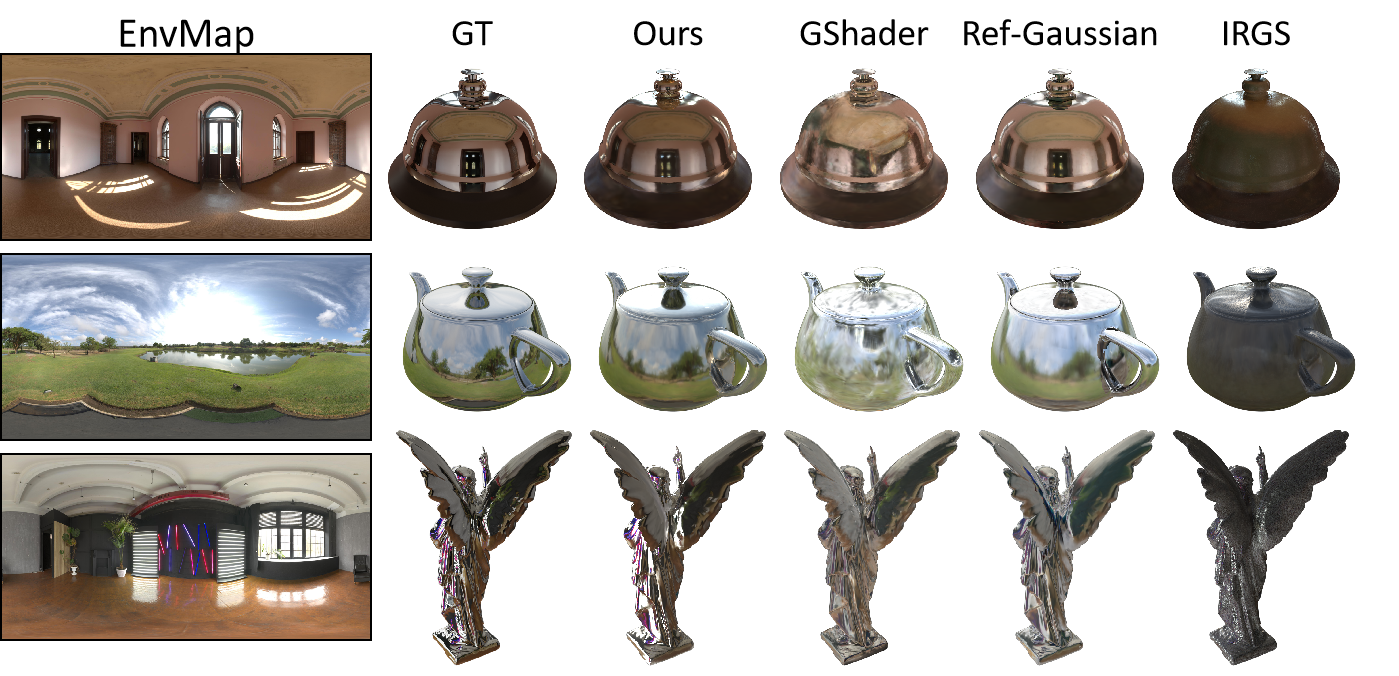}
    \captionsetup{width=0.9\linewidth}
    \caption{Qualitative comparison of relighting results on the Glossy Synthetic dataset~\cite{liu2023nero}.}
    \label{fig:relight_compa}
    \vspace{-1em}
\end{figure}

\begin{figure*}[h!]
    \vspace{-1em}
    \centering
    \includegraphics[width=0.84\textwidth]{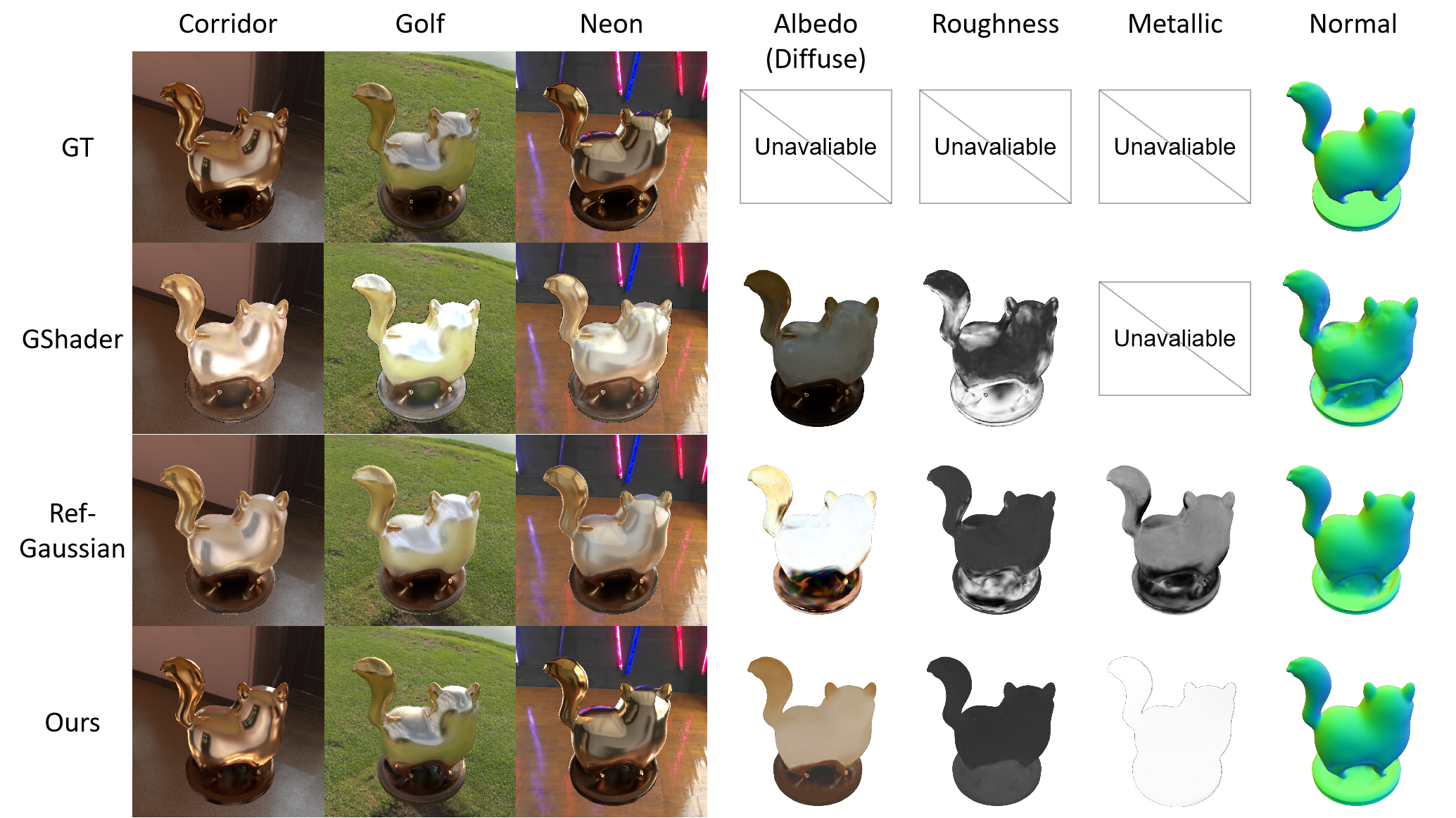}
    \captionsetup{width=0.85\textwidth}
    \caption{Qualitative comparison of normal, material, and lighting estimation, and relighting results (using the Corridor, Golf, and Neon light maps) on the “cat” scene of the Glossy Synthetic Dataset~\cite{liu2023nero}.}
    \label{fig:cat_compa}
    \vspace{-1em}
\end{figure*}

\begin{figure}
    \centering
    \includegraphics[width=0.9\linewidth]{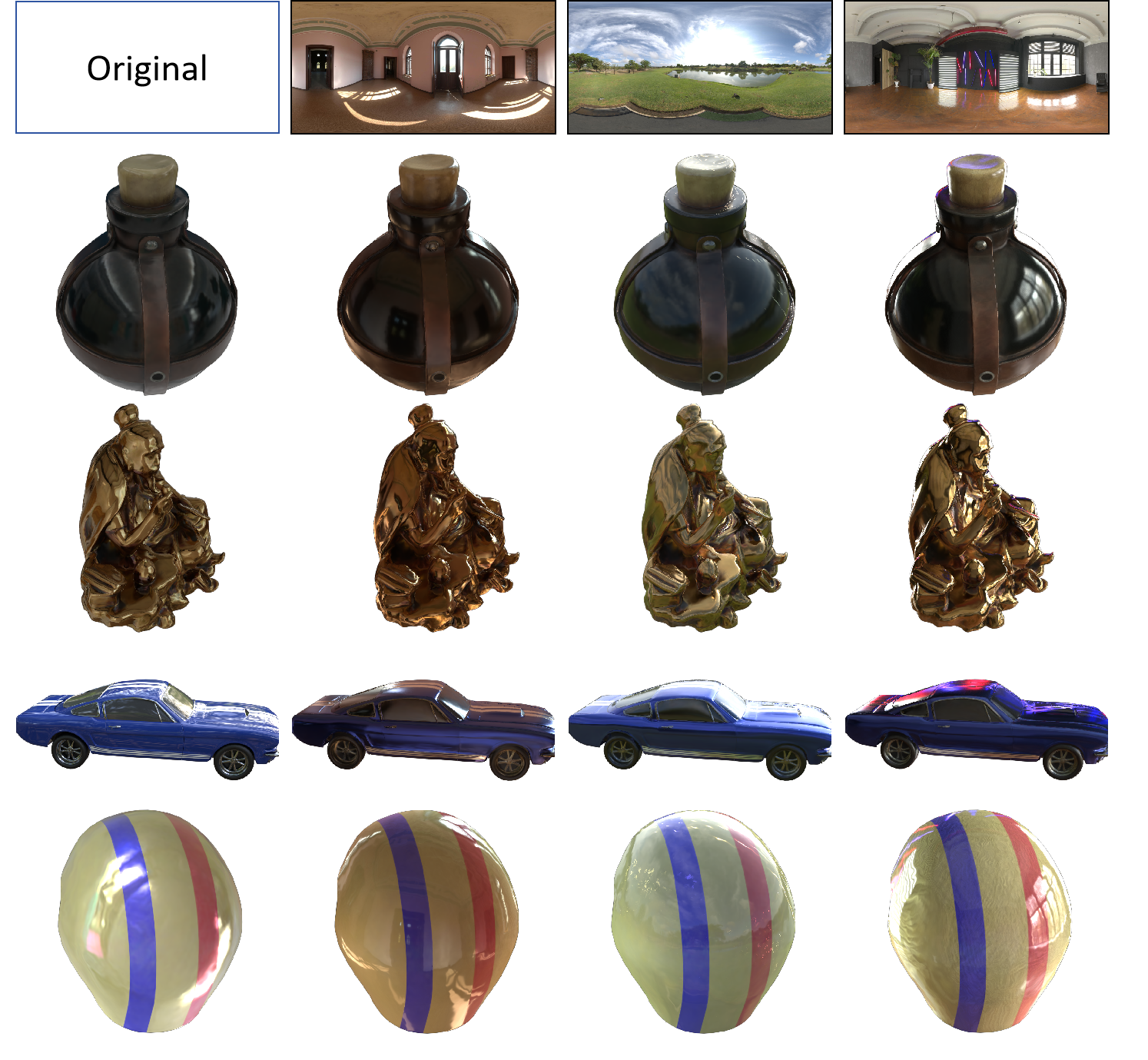}
    \captionsetup{width=0.9\linewidth}
    \caption{Relighting results on the Glossy Synthetic~\cite{liu2023nero} and Shiny Blender dataset~\cite{verbin2022refnerf}.}
    \label{fig:relight_results}
    \vspace{-1em}
\end{figure}
{
\setlength{\parindent}{0pt} 
\textbf{Relighting.}  
Tab.~\ref{tab:relight_psnr} shows our method achieves superior relighting quality on the Glossy Synthetic dataset~\cite{liu2023nero}, outperforming Gaussian-based approaches. 
Qualitative results in Fig.~\ref{fig:relight_compa} demonstrate our inverse rendering produces realistic specular highlights where competitors fail: GShader~\cite{jiang2024gaussianshader} exhibits geometric and material artifacts, IRGS~\cite{gu2025irgs} is limited to diffuse objects, and Ref-Gaussian's simplified rendering equation~\cite{yao2024refgaussian}  yields inaccurate material decomposition despite strong geometry. 
Fig.~\ref{fig:cat_compa} compares our "cat" reconstruction with GS-based methods~\cite{jiang2024gaussianshader, yao2024refgaussian}, showcasing high-fidelity material estimation. 
Additional relighting results for low-metallic objects (Fig.~\ref{fig:relight_results}) further validate the versatility of our method across diverse material types.
}

\subsection{Ablation Study}
In Tab.~\ref{tab:ablation}, we conduct an ablation study to demonstrate the contributions of our method. First, we ablate the geometric prior supervision in the first stage and observe a minor performance drop—this is because the geometric prior supervision primarily acts on inter-reflection regions, which have limited impact on overall performance. Second, we ablate the importance sampling in the inverse rendering stage and notice a substantial performance drop, demonstrating the necessity of importance sampling for smooth object reconstruction and accurate estimation of materials and illumination. Finally, we ablate the specular compensation and once again observe a performance drop, which highlights its benefit to photorealistic reconstruction of glossy objects.

\begin{table}[]
\centering
\captionsetup{width=0.9\linewidth}
\caption{Ablation study on the Glossy Synthetic dataset~\cite{liu2023nero}.}
\label{tab:ablation}
\resizebox{\columnwidth}{!}{%
\begin{tabular}{c|ccc}
\hline
\hline
\multicolumn{1}{l|}{} & \textbf{PSNR↑} & \textbf{SSIM↑} & \textbf{LPIPS↓} \\ \hline
w/o geometry prior & 29.36 & 0.947 & 0.061 \\
w/o important sampling & 28.09 & 0.930 & 0.080 \\
w/o specular compensation & 28.58 & 0.939 & 0.067 \\
Ours & \textbf{29.38} & \textbf{0.947} & \textbf{0.061} \\ \hline \hline
\end{tabular}%
}
\vspace{-1em}
\end{table}
\section{Conclusion}
\label{sec:conc}
We present GOGS, a geometry prior-guided Gaussian surfel framework addressing geometry-material ambiguities in inverse rendering of glossy objects. Our two-stage approach first establishes robust geometry reconstruction under specular interference by integrating foundation model priors with physics-based rendering using split-sum approximation. Subsequently, we perform physically-based material decomposition via Monte Carlo importance sampling of the full rendering equation, leveraging differentiable 2D Gaussian ray tracing. Additionally, a spherical mipmap-based directional encoding mechanism adaptively compensates for high-frequency specular details. Extensive experiments demonstrate state-of-the-art performance in geometry accuracy, material separation, and photorealistic relighting under novel illumination.
{
    \vspace{-2em}
    \bibliographystyle{unsrt}
    \bibliography{main}
}
\appendix
\renewcommand{\thesection}{\Alph{section}}
\clearpage
\maketitlesupplementary
This supplementary document complements the main paper by providing extended qualitative and quantitative analyses to further demonstrate the capabilities of our proposed GOGS framework. Specifically, it expands on the experimental results presented in Sec.~\textcolor{red}{5} of the main paper, including comprehensive visualizations of decomposed model outputs across diverse reflective objects, comparative assessments of recovered environment illumination maps, and extensive relighting scenarios under novel illumination conditions. Additionally, we discuss current methodological limitations and future work.
\section{Foundation Models}
\begin{figure}[b!]
    \centering
    \includegraphics[width=0.9\linewidth]{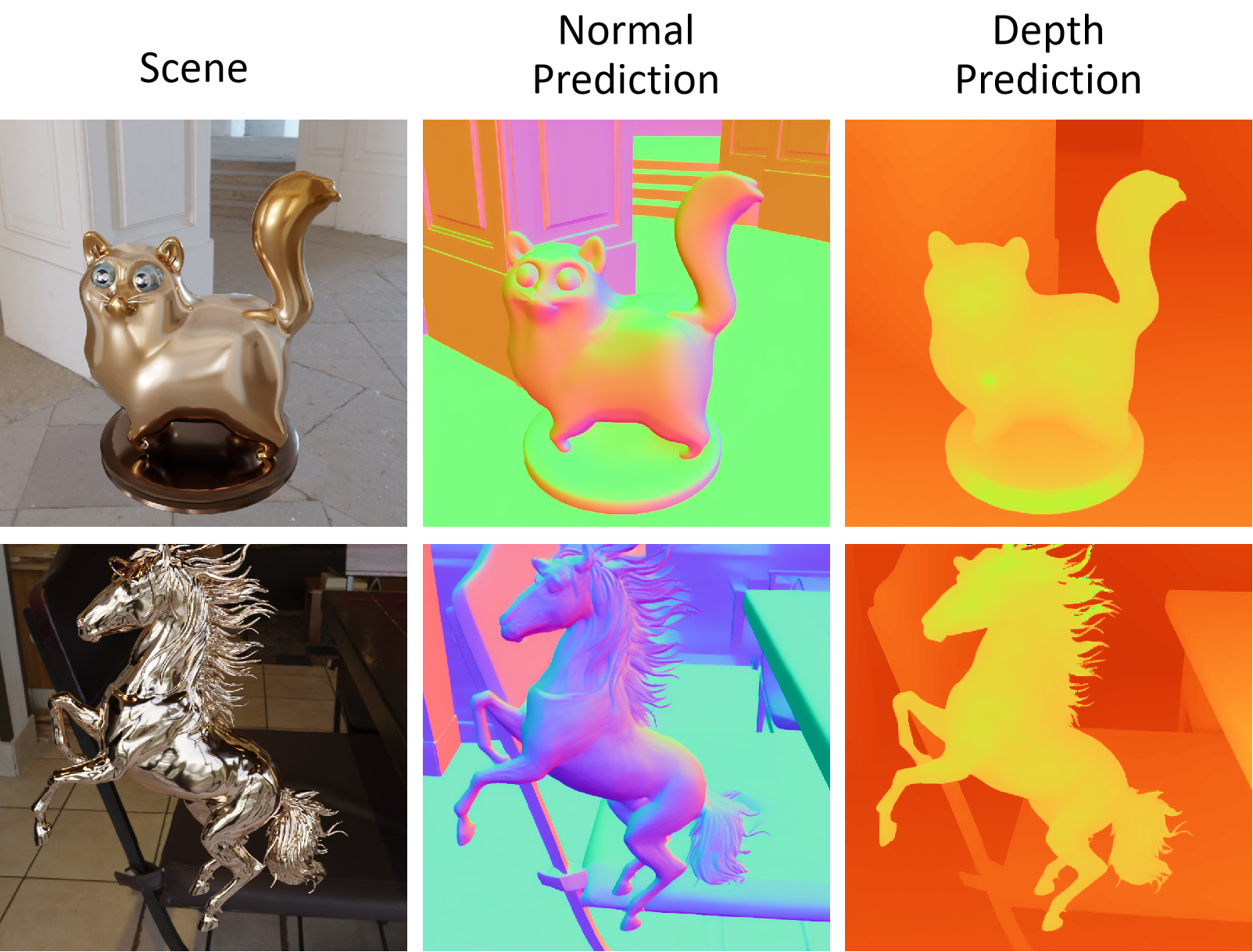}
    \captionsetup{width=0.9\linewidth}
    \caption{Foundation models generate precise depth and normal predictions resilient to specular interference. }
    \label{fig:predictions}
\end{figure}

To mitigate geometric ambiguities under specular surfaces, we leverage foundation models~\cite{Ke_Obukhov_Huang_Metzger_Daudt_Schindler_2023, garcia2025fine} trained on large-scale datasets for monocular depth and normal estimation. These models generate robust geometric predictions by inferring scene geometry from single-view images through learned priors, bypassing the need for multi-view consistency. As demonstrated in Fig.~\ref{fig:predictions}, the predicted depth and normal maps accurately capture surface details across diverse materials including highly reflective regions. The predictions remain robust to complex specular interference and provide reliable pseudo-ground truth for geometric supervision. This capability stems from foundation models' exposure to massive real-world variations during pretraining, enabling them to disambiguate lighting-surface ambiguities that degrade traditional multi-view reconstruction.

It is worth noting that two key factors introduce non-determinism. First, depth and normal priors from external foundation models exhibit inherent randomness during inference, leading to minor variations in predicted geometric cues across repeated runs; second, the depth loss (Eq.~\textcolor{red}{9}), which solves for scale $\omega$ and shift $b$ via least squares optimization—is highly sensitive to minimal distributional discrepancies between rendered depth and priors, and such discrepancies can yield non-unique solutions for $\omega$ and $b$, causing fluctuations in alignment parameters during each iteration. These small variations alter the direction of gradient updates, which amplify over time and ultimately result in differences in the final geometric reconstructions.

\begin{figure}[b!]
    \centering
    \includegraphics[width=1\linewidth]{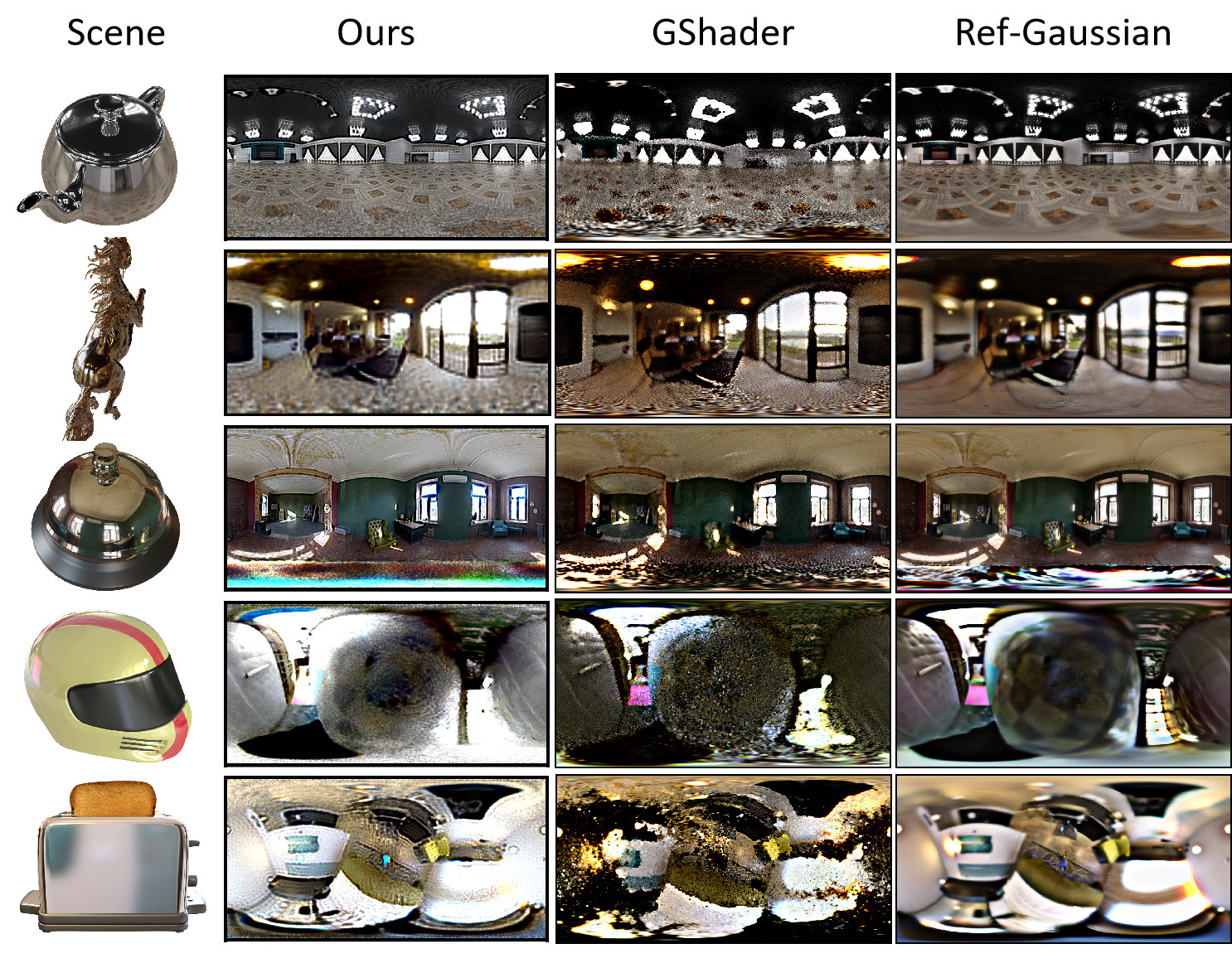}
    \captionsetup{width=0.9\linewidth}
    \caption{Estimated environment maps for reflective object scenes. Our method consistently produces sharper, more artifact-free maps with clearer environmental details compared to competing approaches.}
    \label{fig:envmap_compa_more}
\end{figure}

\begin{figure*}[h!]
    \centering
    \includegraphics[width=1\linewidth]{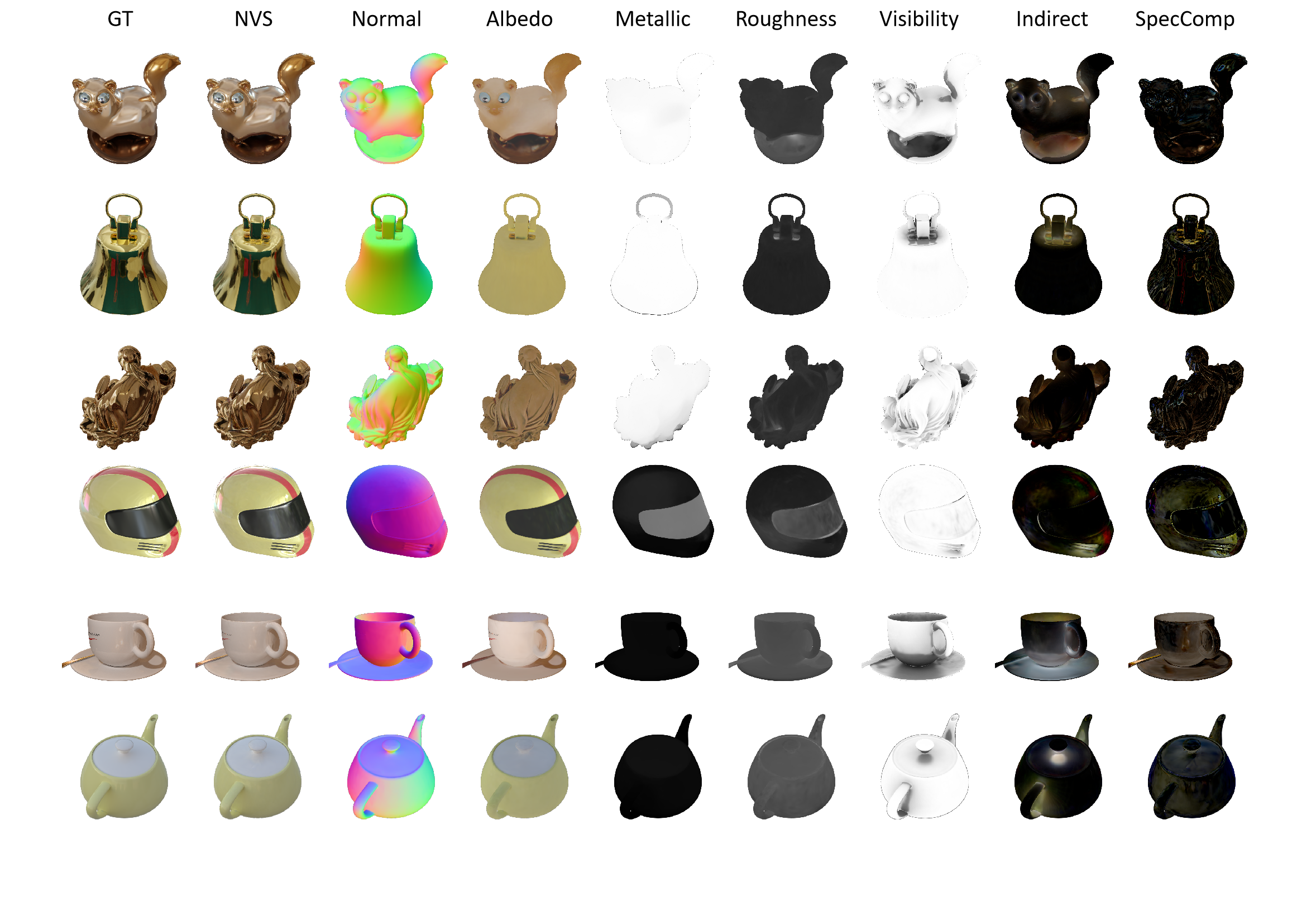}
    \captionsetup{width=0.9\linewidth}
    \caption{Outputs of our model including ground truth (GT), novel view synthesis (NVS), surface normals (visualized with pseudo-colors), albedo, metallic, roughness, visibility, indirect illumination, and specular compensation (SpecComp).}
    \label{fig:results_more}
\end{figure*}

\section{Comparison}
{
\setlength{\parindent}{0pt} 
\textbf{Model Outputs.}  
In Fig.~\ref{fig:results_more}, we present detailed output decompositions of our model applied to reflective objects. From left to right, we show the ground truth (GT) for comparison, our model’s novel view synthesis (NVS) results, pseudo-colored surface normals (for enhanced geometric clarity), albedo (diffuse color), metallic, roughness, visibility maps (encoding direct illumination access), indirect illumination, and specular compensation (SpecComp) outputs. These results demonstrate our method’s ability to effectively disentangle geometry (normals), material properties (albedo, metallic, roughness), and illumination (visibility, indirect, specular compensation) during inverse rendering. By separating these components, our model enables high-quality reconstruction, photorealistic novel view synthesis, and high-fidelity relighting—critical for handling reflective surfaces where traditional methods often exhibit limitations. The disentangled illumination and material outputs, in particular, allow precise adjustment of lighting conditions without compromising surface appearance, a key advantage for applications like product visualization and virtual reality.
}

\begin{table*}[h!]
\centering
\captionsetup{width=0.9\textwidth}
\caption{Quantitative relighting comparisons on three environment maps (corridor, golf, neon) from Glossy Synthetic~\cite{liu2023nero} using three metrics (PSNR ↑, SSIM ↑~\cite{ssim}, LPIPS ↓~\cite{lpips}), with best results per metric highlighted in red.}
\label{tab:relight_compa}
\resizebox{0.95\textwidth}{!}{%
\begin{tabular}{l|cccc|cccc|cccc}
\hline
\hline
 & \textbf{corridor} & \textbf{golf} & \textbf{neon} & \textbf{avg.} & \textbf{corridor} & \textbf{golf} & \textbf{neon} & \textbf{avg.} & \textbf{corridor} & \textbf{golf} & \textbf{neon} & \textbf{avg.} \\ \hline
 & \multicolumn{4}{c|}{PSNR ↑} & \multicolumn{4}{c|}{SSIM↑} & \multicolumn{4}{c}{LPIPS↓} \\ \hline
GShader\cite{jiang2024gaussianshader} & 19.81 & 18.14 & 19.24 & 19.06 & 0.862 & 0.876 & 0.860 & 0.866 & 0.101 & 0.102 & 0.104 & 0.103 \\
Ref-GS\cite{yao2024refgaussian} & 21.83 & 21.69 & 21.36 & 21.63 & 0.900 & \cellcolor[HTML]{FD6864}0.920 & 0.894 & 0.905 & 0.075 & \cellcolor[HTML]{FD6864}0.077 & \cellcolor[HTML]{FD6864}0.082 & \cellcolor[HTML]{FD6864}0.078 \\
IRGS\cite{gu2025irgs} & 22.01 & 21.04 & 20.12 & 21.06 & 0.874 & 0.840 & 0.832 & 0.849 & 0.131 & 0.164 & 0.151 & 0.149 \\
Ours & \cellcolor[HTML]{FD6864}25.75 & \cellcolor[HTML]{FD6864}26.01 & \cellcolor[HTML]{FD6864}22.88 & \cellcolor[HTML]{FD6864}24.89 & \cellcolor[HTML]{FD6864}0.925 & 0.914 & \cellcolor[HTML]{FD6864}0.909 & \cellcolor[HTML]{FD6864}0.916 & \cellcolor[HTML]{FD6864}0.070 & 0.092 & 0.084 & 0.082 \\ \hline
\hline
\end{tabular}%
}
\end{table*}

{
\setlength{\parindent}{0pt} 
\textbf{Environment Maps.}  
In Fig.~\ref{fig:envmap_compa_more}, we present a comparison of estimated environment maps for various reflective object scenes (left column), including a teapot, horse, car, and helmet. From left to right after the scene column, we show results from our model, GShader~\cite{jiang2024gaussianshader}, Ref-Gaussian(Ref-GS)~\cite{yao2024refgaussian}. Our method effectively disentangles illumination from scene geometry and material properties, resulting in sharper environment maps with fewer artifacts—such as the crisp ceiling patterns in the teapot scene, distinct window frames in the horse scene—compared to GShader’s slightly blurred outputs~\cite{jiang2024gaussianshader} and Ref-Gaussian’s muted details~\cite{yao2024refgaussian}. Notably, these methods exhibit severe noise and color distortion in most cases (e.g., the car and helmet scenes), while our model maintains consistent clarity across all scenes. These results demonstrate our method’s superior ability to recover high-fidelity environment illumination, which is critical for realistic rendering and relighting of reflective surfaces—an essential capability for applications like product visualization and virtual reality.
}

\begin{figure}[]
    \centering
    \includegraphics[width=1\linewidth]{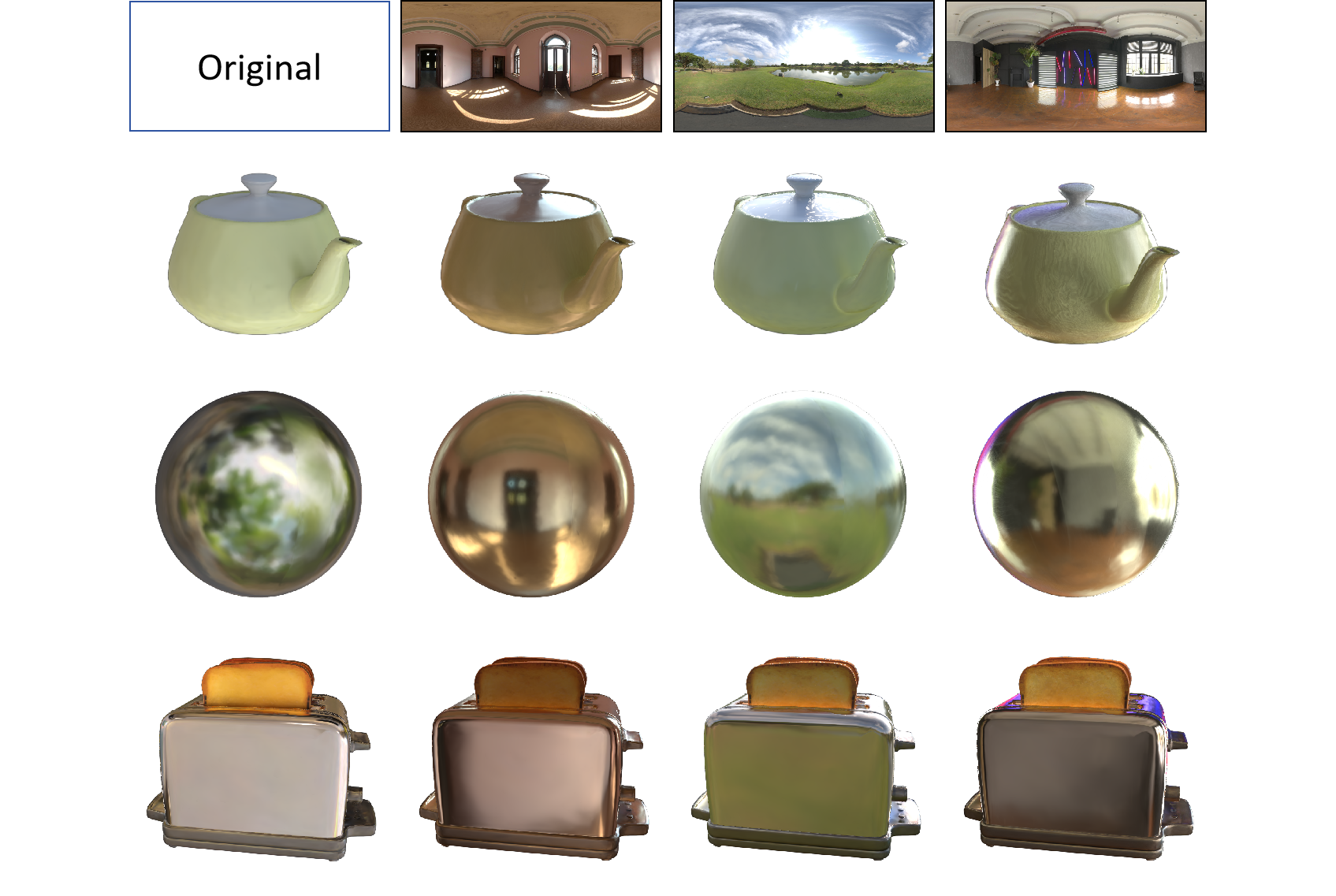}
    \captionsetup{width=0.9\linewidth}
    \caption{Our GOGS framework generates high-fidelity relighting results across diverse objects, consistently preserving geometric fidelity and material properties while producing sharp, physically plausible specular reflections under novel illumination conditions.}
    \label{fig:relight_results_more}
\end{figure}

{
\setlength{\parindent}{0pt} 
\textbf{Relighting.}  
In Fig.~\ref{fig:relight_results_more}, we demonstrate GOGS's relighting capabilities across surfaces with diverse reflectance properties. For each exemplar object, we preserve the original rendering under default illumination (leftmost column), followed by relighting under three novel illumination conditions (subsequent columns). Our method consistently disentangles illumination from intrinsic scene properties, maintaining geometric fidelity (e.g., fine structural details) and material characteristics across lighting variations. Relighted outputs exhibit physically accurate reflections: highly specular surfaces precisely mirror environmental features, while matte surfaces retain color/texture consistency.These results validate our model's capability for high-fidelity relighting of arbitrary materials under varied illuminations.
}

Tab.~\ref{tab:relight_compa} summarizes the relighting performance of four methods (GShader~\cite{jiang2024gaussianshader}, Ref-GS~\cite{yao2024refgaussian}, IRGS~\cite{gu2025irgs}, and ours) on three representative environment maps (corridor, golf, neon) using three widely adopted metrics: PSNR↑, SSIM↑~\cite{ssim}, and LPIPS↓~\cite{lpips}. For each metric, we report results on individual environment maps and their average (avg.), with the best results highlighted in red.  
Consistently, our method achieves the highest average PSNR and SSIM across all environment maps, outperforming the second-best method by significant margins. These results validate the effectiveness of our method in disentangling illumination from scene geometry and material properties, which is critical for high-fidelity relighting applications like product visualization and virtual reality.
These results validate our model's capability for high-fidelity relighting of arbitrary materials under varied illuminations.

\section{Limitation}
Our framework has certain limitations requiring further investigation. While optimized for specular reflections, our Monte Carlo importance sampling exhibits suboptimal performance on complex mixed-material surfaces due to differing BRDF characteristics. To address this, future work will focus on adaptive importance sampling techniques capable of handling complex mixed materials and generalizing across material roughness. Additionally, computational costs from our ray tracing implementation and spherical encoding currently preclude real-time applications, despite offering significant speed improvements over NeRF-based methods. Tab.~\ref{tab:time} quantifies this trade-off: our model achieves a training time of 1.33 hours without specular compensation—far faster than NeRF-based ENVIDR (5.84 hours)—but slower than lightweight models like 3DGS-DR (0.35 hours) and GShader (0.48 hours). This aligns with our observation that while we outperform NeRF-based methods, real-time capability remains a challenge. To address this, we plan to accelerate rendering by baking incident radiance through precomputed radiance transfer schemes while maintaining physical accuracy.Finally, while our 2D Gaussian ray tracing approximates visibility and indirect illumination, it remains insufficient for capturing high-fidelity multi-bounce specular inter-reflections. Real-world light transport in glossy surfaces involves theoretically infinite path bounces—a phenomenon prohibitively expensive to simulate using path-tracing in inverse rendering due to its computational burden. To address this limitation, we will explore novel neural network-based approximation techniques for modeling complex light paths.

\begin{table}[]
\centering
\caption{Quantitative comparison of training time efficiency on the Shiny Blender dataset~\cite{verbin2022refnerf}.}
\label{tab:time}
\resizebox{\columnwidth}{!}{%
\begin{tabular}{c|ccccc}
\hline
\hline
\Large
 & \multicolumn{5}{c}{Model} \\ \cline{2-6} 
\multirow{-2}{*}{\begin{tabular}[c]{@{}c@{}}Efficiency \\ evaluation\end{tabular}} & ENVIDR\cite{liang2023envidr} & 3DGS-DR\cite{ye20243dgsdr} & Ref-GS\cite{yao2024refgaussian} & IRGS\cite{gu2025irgs} & Ours \\ \hline
\begin{tabular}[c]{@{}c@{}}Training \\ time (h)↓\end{tabular} & 5.84 & \cellcolor[HTML]{FD6864}0.35 & 0.58 & 0.70 & 1.33 \\ \hline
\hline
\end{tabular}%
}
\end{table}

\end{document}